# On the Generation of Alternative Explanations with Implications for Belief Revision

Eugene Santos Jr.
Department of Computer Science
Brown University
Providence, RI 02912

## Abstract

In general, the *best* explanation for a given observation makes no promises on how good it is with respect to other alternative explanations. A major deficiency of message-passing schemes for belief revision in Bayesian networks is their inability to generate alternatives beyond the second best. In this paper, we present a general approach based on linear constraint systems that naturally generates alternative explanations in an orderly and highly efficient manner. This approach is then applied to cost-based abduction problems as well as belief revision in Bayesian networks.

## 1 INTRODUCTION

We are constantly faced with the problem of explaining the observations we have gathered with our senses. Our explanations are constructed by assuming certain facts or hypotheses which support our observations. For example, suppose I decide to phone my friend Tony at the office. After several rings, no one has answered the phone. From this, I conclude that Tony is not at the office. Our observation in this case is that no one answered the phone. Our explanation for this is that Tony is not at the office. The reasoning process we have just used is called *abductive explanation* (Charniak & Shimony [1990]; Hobbs et al. [1988]; Peng & Reggia [1990]; Selman & Levesque [1990]; Shanahan [1989]). It is often formalized as the process of finding certain hypotheses which can explain or prove the things we observe.

Although we used the word "conclude" in our story, our confidence in our solution may not be absolute. Suppose that I also know for a fact that Tony sometimes disconnects the phone to take a nap in the office. Now, I have an alternative explanation for why the phone was not answered. In general, there are many possible explanations for any given observation, but yet, we often express confidence in one explanation over the others and choose it to be our solution. It is this fact that distinguishes abductive reasoning from deductive reasoning.

In current approaches to modeling abduction, confidence in an explanation is defined by some measure on the set of hypotheses it represents. Such measures include minimal cardinality (Genesereth [1984]; Kautz & Allen [1986]), parsimonious covering theory (Peng & Reggia [1990]), most-probable explanation (Pearl [1988]) and minimal cost proofs (Charniak & Shimony [1990]; Hobbs et al. [1988]; Stickel [1988]). These approaches provide us with a model for choosing a "best" explanation.

In particular, we are interested in *minimal cost proofs* found in the *cost-based abduction model* (Charniak & Shimony [1990]).[1] Under this model, *costs* are associated with individual hypotheses. The use of a hypothesis in an explanation incurs the cost associated with the hypothesis. Thus the cost of an explanation is simply the sum of the costs of the individual hypotheses used. These costs now represent our confidence in each explanation and establishes an ordering on the explanations.

Since cost-based abduction has been shown to be an NP-Hard problem (Charniak & Shimony [1990]), the runtime of standard searching techniques grows exponentially with the size of the problem. In (Santos [1991a]; Santos [1991b]; Santos [1991c]), it was shown that any cost-based abduction problem may be transformed into an equivalent *linear constraint satisfaction problem*, and the latter may be solved by utilizing the highly efficient optimization tools of operations research. Indeed, empirical studies in (Santos [1991a]; Santos [1991c]) showed that the approach is computationally practical and superior to search style techniques. Our linear constraint satisfaction approach actually exhibited a subexponential growth rate.

---

[1] Cost-based abduction is a minor variant of *weighted abduction* (Hobbs et al. [1988]; Stickel [1988])

Now suppose that we further know that my friend Tony spends nearly all of his time in the office working, sleeping, and eating. This knowledge will significantly increase the likelihood of the phone being disconnected as an alternative explanation. Even though our measures may still choose the initial explanation, they in general make no promises on how good this choice is with respect to our alternative. This issue is especially important in domains such as medical diagnosis where careful consideration of alternative diagnoses/explanations is necessary. Thus, the ability to generate alternative explanations should exist in any complete model of abductive reasoning.

In this regard, a major deficiency of message-passing schemes (Pearl [1988]) for belief revision in Bayesian networks is its inability to generate alternative explanations in an ordered manner beyond the second best. By considering the equivalent problem in terms of constraint systems, we can generate the consecutive next best explanations. In this paper, we present an approach based on our linear constraint systems to generate alternative explanations in order of cost.

In Section 2, we present an overview of constraint systems and cost-based abduction. In Section 3, we present our approach to generating alternative explanations. In Section 4, we consider how our constraint systems may be applied to belief revision in Bayesian networks. Finally, in Section 5, we conclude our discussion and give some final thoughts concerning alternative explanations.

## 2   CONSTRAINT SYSTEMS

We now present a brief overview of the formulation of cost-based abduction problems as constraint systems. Details and complete proofs can be found in (Santos [1991a]; Santos [1991c]).

NOTATION. $\Re$ denotes the set of real numbers.

DEFINITION 2.1. *A* WAODAG *(or weighted* AND/OR *directed acyclic graph)*[2] *is a 4-tuple* $(G, c, r, S)$, *where:*

1. $G$ *is a directed acyclic graph, $G = (V, E)$.*

2. $c$ *is a function from $V \times \{\text{true}, \text{false}\}$ to $\Re$, called the* cost *function.*

3. $r$ *is a function from $V$ to $\{\text{AND}, \text{OR}\}$, called the* label. *A node labeled* AND *is called an* AND-*node, etc.*

4. *S is a subset of nodes in V called the* evidence *nodes.*

NOTATION. $V_H$ is the subset of nodes with zero indegree called the *hypothesis nodes*.

---

[2]Slight generalization of (Charniak & Shimony [1990]).

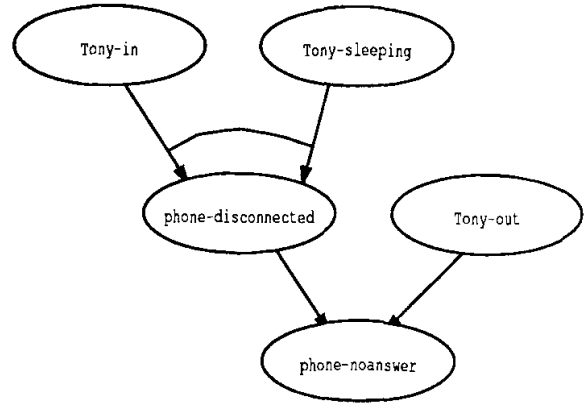

FIG. 2.1. Tony's office habits. phone-disconnected is the only AND-node, phone-noanswer is the only OR-node and the remaining nodes are hypothesis nodes.

DEFINITION 2.2. *A truth assignment for a* WAODAG $W = (G, c, r, S)$ *where* $G = (V, E)$ *is a function $e$ from $V$ to $\{\text{true}, \text{false}\}$. We say that such a function is* valid *iff the following conditions hold:*

1. *For all* AND-*nodes $q$, $e(q) = $ true iff for all nodes $p$ such that $(p, q)$ is an edge in $E$, $e(p) = $ true.*

2. *For all* OR-*nodes $q$, $e(q) = $ true iff there exists a node $p$ such that $(p, q)$ is an edge in $E$ and $e(p) = $ true.*

*Furthermore, we say that $e$ is an* explanation *iff $e$ is valid and for each node $q$ in $S$, $e(q) = $ true.*

DEFINITION 2.3. *We define the* cost *of an explanation $e$ for $W = (G, c, r, S)$ where $G = (V, E)$ as*

$$C(e) = \sum_{q \in V} c(q, e(q)).$$

*An explanation $e$ which minimizes $C$ is called a* best explanation *for $W$.*

Consider the WAODAG representing the situation with our friend Tony (see Figure 2.1). We first assume that there is no cost for assigning a node to false. Next, assume that assigning Tony-in, Tony-sleeping and Tony-out to true have costs 5, 4 and 8, respectively, and that the costs of assigning true to all non-hypothesis nodes is zero. The minimal cost proof for this WAODAG is the hypotheses set {Tony-out} with a cost of 8.

We now define constraint systems as follows:

NOTATION. For each node $q$ in $V$, let $D_q = \{p | (p, q)$ is an edge in $E\}$, the parents of $q$. $|D_q|$ is the cardinality of $D_q$.

DEFINITION 2.4. *A constraint system is a 3-tuple $(\Gamma, I, \psi)$ where $\Gamma$ is a finite set of variables, $I$ is a finite set of linear inequalities based on $\Gamma$, and $\psi$ is a* *function from* $\Gamma \times \{\text{true}, \text{false}\}$ *to* $\Re$. *Given a* WAODAG $W = (G, c, r, S)$ *where* $G = (V, E)$, *we can construct a constraint system* $L(W) = (\Gamma, I, \psi)$ *where:*

1. $\Gamma$ *is a set of variables indexed by* $V$, *that is*, $\Gamma = \{x_q | q \in V\}$.
2. $\psi(x_q, X) = c(q, X)$ *for all* $q \in V$ *and* $X \in \{\text{true}, \text{false}\}$.
3. $I$ *is the collection of all inequalities of the forms given below:*

$$x_q \leq x_p \in I \text{ for each } p \in D_q \text{ if } r(q) = \text{AND} \quad (1)$$

$$\sum_{p \in D_q} x_p - |D_q| + 1 \leq x_q \in I \text{ if } r(q) = \text{AND} \quad (2)$$

$$\sum_{p \in D_q} x_p \geq x_q \in I \text{ if } r(q) = \text{OR} \quad (3)$$

$$x_q \geq x_p \in I \text{ for each } p \in D_q \text{ if } r(q) = \text{OR} \quad (4)$$

*We say that* $L(W)$ *is induced by* $W$. *Furthermore, by including the additional constraints:*

$$x_q = 1 \text{ if } q \in S, \quad (5)$$

*we say that the resulting constraint system is induced evidentially by* $W$ *and is denoted by* $L_E(W)$.

DEFINITION 2.5. *A variable assignment for a constraint system* $L = (\Gamma, I, \psi)$ *is a function* $s$ *from* $\Gamma$ *to* $\Re$. *Furthermore,*

1. *If the range of* $s$ *is* $\{0, 1\}$, *then* $s$ *is a 0-1 assignment.*
2. *If* $s$ *satisfies all the constraints in* $I$, *then* $s$ *is a solution for* $L$.
3. *If* $s$ *is a solution for* $L$ *and is a 0-1 assignment, then* $s$ *is a 0-1 solution for* $L$.

Given a 0-1 assignment $s$ for $L(W)$, we can construct a truth assignment $e$ for $W$ as follows:

1. For all $q$ in $V$, $s(x_q) = 1$ iff $e(q) = \text{true}$.
2. For all $q$ in $V$, $s(x_q) = 0$ iff $e(q) = \text{false}$.

Conversely, given a truth assignment $e$ for $W$, we can construct a 0-1 assignment $s$ for $L(W)$.

NOTATION. $e_s$ and $s_e$ denote, respectively, a truth assignment $e$ constructed from a 0-1 assignment $s$, and a 0-1 assignment $s$ constructed from a truth assignment $e$.

We can show that all explanations for a given WAODAG $W$ have corresponding 0-1 solutions for $L_E(W)$ and vice versa.

THEOREM 2.1. *If* $e$ *is an explanation for* $W$, *then* $s_e$ *is a solution of* $L(W)$.

THEOREM 2.2. *If* $s$ *is a 0-1 solution of* $L_E(W)$, *then* $e_s$ *is an explanation for* $W$.

It follows from Theorems 2.1 and 2.2 that 0-1 solutions for constraint systems are the counterparts of explanations for WAODAGs. Thus, by augmenting a WAODAG induced constraint system with a cost function, the notion of the cost of an explanation for a WAODAG can be transformed into the notion of the cost of a 0-1 solution for the constraint system.

DEFINITION 2.6. *Given a constraint system* $L = (\Gamma, I, \psi)$, *we construct a function* $\Theta_L$ *from variable assignments to* $\Re$ *as follows:*

$$\Theta_L(s) = \sum_{x_q \in \Gamma} \{s(x_q)\psi(x_q, \text{true}) + (1 - s(x_q))\psi(x_q, \text{false})\}.$$

$\Theta_L$ *is called the* objective function *of* $L$.

DEFINITION 2.7. *An* optimal 0-1 solution *for a constraint system* $L = (\Gamma, I, \psi)$ *is a 0-1 solution which minimizes* $\Theta_L$.

Clearly, Definition 2.6 is identical to Definition 2.3. Thus, it follows from Theorems 2.1 and 2.2 and the relationship between node assignments and variable assignments that an optimal 0-1 solution in $L_E(W)$ is a best explanation for $W$ and vice versa.

As we observed in (Santos [1991a]; Santos [1991c]), $I$ and $\Theta_L$ are the elements of a *linear program* in operations research (Nemhauser, Kan & Todd [1989]). Extremely efficient and practical optimization techniques such as the *Simplex method* and *Karmarkar's projective scaling algorithm* (Nemhauser, Kan & Todd [1989]) are available for use in minimizing $\Theta_L$ with respect to the constraints in $I$.

Although solving the linear program was sufficient to obtain an optimal 0-1 solution for most of our test problems in (Santos [1991a]; Santos [1991c]), it was sometimes necessary to employ a *branch and bound* technique using the linear program to compute lower bounds. Complete details concerning the branch and bound algorithm can be found in (Santos [1991a]; Santos [1991c]). This technique enables us to avoid searching through all possible solutions by utilizing the lower bounds computed by the linear program as a guide. Experiments performed in (Santos [1991a]; Santos [1991c]) shown the practicality and efficiency of this approach for solving cost-based abduction problems. Also, it can be applied to any constraint system regardless of whether or not they are WAODAG induced.

## 3 GENERATING ALTERNATIVE EXPLANATIONS

In abductive explanation, having alternative explanations is often useful and sometimes necessary. Having the 2nd best, 3rd best, and so on, can provide a useful gauge on the quality of the best explanation. In this section, we present techniques for extracting alternative explanations in order of their associated costs.

To generate the alternative explanations, we solve a sequence of constraint systems. This sequence consists of constraint systems each of which are derived from the constraint systems earlier in the sequence. The initial constraint system is the original constraint system which determines the first optimal solution. The subsequent constraint systems are generated using the following schema: Consider $L_1 = (\Gamma, I_1, \psi)$, our initial constraint system. Let $s_1$ be the optimal 0-1 solution of $L_1$. We define a new problem $L_2$ as the successor of $L_1$. $L_2$ is identical to $L_1$ except for the additional constraint

$$\sum_{x_q \in \Gamma} F(s_1, x_q) \leq |\Gamma| - 1$$

where for each $x_q \in \Gamma$,

$$F(s_1, x_q) = \begin{cases} x_q & \text{if } s_1(x_q) = 1 \\ (1 - x_q) & \text{if } s_1(x_q) = 0 \end{cases}$$

Note that the new problem does not have $s_1$ as its optimal 0-1 solution since the variable assignment would violate the new constraint.

Let $s_2$ be the optimal 0-1 solution, if any, to $L_2$. This will be the second best 0-1 solution. To continue the search for the next best explanation, we simply define a successor to the last constraint system, in this case, $L_2$. When the current constraint system does not yield any solution, all possible explanations have been generated and we are finished.

ALGORITHM 3.1. *Given a constraint system $L = (\Gamma, I, \psi)$, generate all the 0-1 solutions for $L$ in order of cost.*

1. *(Initialization) Set $I_1 := I$, $L_1 := (\Gamma, I_1, \psi)$ and $k := 1$.*

2. *Compute the optimal 0-1 solution for $L_k$. If there is no feasible solution, then go to step 7. Otherwise, let $s_k$ be the solution.*

3. *$k := k + 1$.*

4. *Let $I_k := I_{k-1} \cup c_{k-1}$ where $c_{k-1}$ contains the single constraint*

$$\sum_{x_q \in \Gamma} F(s_{k-1}, x_q) \leq |\Gamma| - 1 \qquad (6)$$

*where for each $x_q \in \Gamma$,*

$$F(s_{k-1}, x_q) = \begin{cases} x_q & \text{if } s_{k-1}(x_q) = 1 \\ (1 - x_q) & \text{if } s_{k-1}(x_q) = 0 \end{cases}$$

5. *Let $L_k := (\Gamma, I_k, \psi)$.*

6. *Go to step 2.*

7. *(Solutions) Print $s_1, s_2, \ldots, s_{k-1}$.*

The method we have just described can be classified as a *cutting plane method* in operations research (Nemhauser, Kan & Todd [1989]). Since each derived constraint system differs only in an additional constraint from some previously solved problem, efficient incremental techniques such as the *dual simplex method* can be applied here in a fashion similar to the one which is used in the branch and bound algorithm.

THEOREM 3.1. *Constraint system $L_n$ in Algorithm 3.1 determines the n-th best 0-1 solution for $L$.*

The algorithm we have just presented can be applied to any constraint system. However, there are certain situations where generating all possible explanations may not be particularly desirable. Returning to our friend Tony above, consider the following additional information: Tony is as likely to be awake as be asleep at any time since he can always get to sleep in any environment. This implies that for the hypothesis that Tony is awake, the difference in the cost of being true and it being false is 0. If we look at our original explanation that Tony is not in the office, we must augment it with our guess as to whether he is asleep or not. With our assumptions, there is no way to choose between asleep and awake. However, since Tony is not in the office, the hypothesis involving his consciousness has no impact towards explaining the observation (see Figure 2.1).

If the algorithm first chooses that Tony is asleep, then the next alternative would be the same set of assignments except for Tony being awake. However, this new alternative explanation is uninteresting. In general, it may be the case that we may run into an overly large number of these types of uninteresting explanations. We now proceed to present an approach to deal with this problem.

DEFINITION 3.1. *Given a WAODAG $W = (G, c, r, S)$ where $G = (V, E)$ and $H \subseteq V_H$, an explanation $e$ for $W$ is said to be* consistent with $H$ *iff for all $h$ in $H$, $e(h) = \text{true}$. The base set $H(e)$ of $e$ is the subset of $V_H$ consisting of all $h$ in $V_H$ where $e(h) = \text{true}$.*

In WAODAGs, finding the best explanation is tantamount to finding the best set of hypotheses we need to assume.

DEFINITION 3.2. *The support-set $K(e)$ of an explanation $e$ is the set consisting of all nodes $m$ in $V$ such that $e(m) = \text{true}$.*

PROPOSITION 3.2. *For every explanation $e$ for $W$, $H(e) = K(e) \cap V_H$.*

The following propositions follow immediately from the properties of WAODAGs:

PROPOSITION 3.3. *Let $e_1$ and $e_2$ be explanations for $W$. $H(e_1) = H(e_2)$ iff $K(e_1) = K(e_2)$.*

PROPOSITION 3.4. *Let $e$ be an explanation for $W$. For each $H(e) \subseteq H \subseteq V_H$, there exists an explanation $e'$ for $W$ such that $H(e') = H$.*

THEOREM 3.5. *Let $e_1$ and $e_2$ be explanations for $W$.*

1. $H(e_1) \subseteq H(e_2)$ iff $K(e_1) \subseteq K(e_2)$.
2. $H(e_1) \subset H(e_2)$ iff $K(e_1) \subset K(e_2)$.

THEOREM 3.6. *There exists a 1-1 and onto mapping between $2^{V_H}$ and the set of all possible truth assignments for $W$.*

THEOREM 3.7. *If $e$ is an explanation for $W$, then there exists at least $2^{|V_H - H(e)|}$ explanations for $W$ which are consistent with $H(e)$.*

In general, we see that there are an exponential number of explanations for a given WAODAG. However, from Theorem 3.7, it seems that the majority of these explanations are formed from a possibly small number of "simpler" and more interesting explanations which utilize smaller numbers of hypotheses. The following question naturally arises: Do these additional explanations provide any new or important information?

DEFINITION 3.3. *A WAODAG $W$ is* monotonic *iff for every two explanations $e_1$ and $e_2$ for $W$, $K(e_1) \subseteq K(e_2)$ implies $C(e_1) \leq C(e_2)$. $W$ is* strictly monotonic *iff $W$ is monotonic, and for every two explanations $e_1$ and $e_2$ for $W$, $K(e_1) \subset K(e_2)$ implies $C(e_1) < C(e_2)$.*

PROPOSITION 3.8. *If $c(v, \text{true}) \geq c(v, \text{false})$ for all $v$ in $V$, then $W$ is monotonic. If $c(v, \text{true}) > c(v, \text{false})$ for all $v$ in $V$, then $W$ is strictly monotonic.*

THEOREM 3.9. *A WAODAG $W$ is monotonic iff for every two explanations $e_1$ and $e_2$ for $W$, $H(e_1) \subseteq H(e_2)$ implies $C(e_1) \leq C(e_2)$. $W$ is strictly monotonic iff $W$ is monotonic, and for every two explanations $e_1$ and $e_2$ for $W$, $H(e_1) \subset H(e_2)$ implies $C(e_1) < C(e_2)$.*

Proposition 3.8 and Theorem 3.9 together show that in a monotonic WAODAG, "simpler" explanations are preferred due to the lower associated costs. The assumption of monotonicity is reasonable in many cases as pointed out by (Charniak & Shimony [1990]) and characterized in (Charniak & Goldman [1988]). Our goal is to generate these explanations in order of cost without having to consider the remaining exponential number of explanations.

DEFINITION 3.4. *$e$ is* cardinal *iff there are no explanations $e'$ such that $H(e') \subset H(e)$.*

Intuitively, a cardinal explanation is among the "simplest" of explanations we wish to consider.

THEOREM 3.10. *If $W$ is strictly monotonic, then any best explanation for $W$ is cardinal.*

All the definitions given above involving WAODAGs can be carried over to WAODAG induced constraint systems.

Similar to Algorithm 3.1, the best cardinal explanation, 2nd best, 3rd best, etc. may be generated by constructing a sequence of constraint systems $L_1, L_2, \ldots$. Instead of introducing the additional constraint (6) to $L_k$, we introduce

$$\sum_{x_q \in H(s_{k-1})} x_q \leq |H(s_{k-1})| - 1.$$

LEMMA 3.11. *Let $W$ be strictly monotonic. If $s_n$ is the optimal 0-1 solution for the constraint system $L_n$, then $s_n$ is a cardinal 0-1 solution for $L$.*

THEOREM 3.12. *Let $W$ be strictly monotonic. The constraint system $L_n$ determines the n-th best cardinal 0-1 solution.*

Our notion of cardinal explanations is very similar to the notion of *irredundancy* found in *parsimonious covering theory* for modeling medical diagnosis (Peng & Reggia [1990]). A *diagnostic problem* (Peng & Reggia [1990]) is a two-layer network consisting of a layer of *manifestations* which are causally affected by a layer of *disorders*. Given a subset of the manifestations as evidence, a subset of disorders must be chosen to best explain the manifestations based on parsimonious covering theory. A collection of disorders which can explain the manifestations is called a *cover*. A cover is said to be irredundant if none of its proper subsets is also a cover.

A limitation of parsimonious covering theory as pointed out by Peng and Reggia (Peng & Reggia [1990]) is the large number of covers which are considered "best". In order to further select from these potential explanations, some additional criteria must be used. Basic parsimonious covering theory is extended to incorporate probability theory. The potential of an explanation is now measured by some probability. With the addition of probabilities, care must be taken in choosing which covers are to be inspected. For example, consider the following analogous problem in cost-based abduction: A set of disorders $D$ can adequately explain manifestations $M$. Let $d$ be a fairly common disorder which explains manifestation $m$. Assume $d$ is not in $D$ but $m$ is present in $M$. Furthermore, assume $c(d, F) > c(d, T)$. Thus, $D \cup \{d\}$ is a better explanation than $D$, despite the fact that $D \cup \{d\}$ is a superset of $D$.

Although this modified algorithm works only for $W$ being strictly monotonic, we can modify any non-strictly monotonic problem to make it applicable. In essence, the strict monotonicity simply implies that we should always have a preference for a false assignment over a true assignment. By introducing an arbitrarily small positive difference between the cost for true and the cost for false in the original problem, we can now determine the cardinal solutions of the new problem which turns out to be identical to those of the original.

## 4 BAYESIAN NETWORKS

*Bayesian networks* have become an important tool in modeling probabilistic reasoning. The inherent representational power of these networks provides a very promising approach. In particular, *belief revision* in Bayesian networks is the process of finding the best interpretation for some given piece of evidence. This, of course, is a cornerstone of abductive explanation.

Since we are interested in abduction, existing effective algorithms for belief revision should be considered. One such algorithm is given by Pearl in (Pearl [1988]) which is based on a *message passing scheme*. However, except for simple networks such as polytrees, the method is rather complicated to apply. Also, as Pearl points out in Chapter 5 in (Pearl [1988]), this algorithm cannot guarantee the generation of alternative explanations beyond the second best.

Our goal in this section is to apply our linear constraint satisfaction approach to Bayesian networks. This entails constructing a constraint system which is computationally equivalent to the Bayesian network. Although this could be done by first transforming the Bayesian network into a cost-based abduction graph (Charniak & Shimony [1990]) and then transforming the graph into a constraint system (Santos [1991a]; Santos [1991c]), a more natural and straightforward method will be given below. We will show how to directly transform a Bayesian network into an equivalent constraint system.

We first observe that a Bayesian network can be completely described by a finite collection of random variables (or simply, r.v.s) and a finite set of conditional probabilities based on the r.v.s.[3]

NOTATION. Throughout the remainder of this paper, upper case italicized letters such as $A, B, \ldots$ will represent r.v.s and lower case italicized letters such as $a, b, \ldots$ will represent the possible assignments to the associated upper case letter r.v., in this case, $A, B, \ldots$. Subscripted upper case letters which are not italicized are variables in a constraint system which explicitly represent the instantiation of the associated r.v. with the item in the subscript. For example, $A_a$ denotes the instantiation of r.v. $A$ with value $a$.

NOTATION. Given a r.v. $A$, the set of possible values for $A$ called the *range* of $A$ will be denoted by $R(A)$.

Given a Bayesian network, we can construct an ordered pair $(V, P)$ where $V$ is the set of r.v.s in the network and $P$ is a set of conditional probabilities associated with the network. $P(A = a | C_1 = c_1, \ldots, C_n = c_n) \in P$ iff $C_1, \ldots, C_n$ are all the immediate parents of $A$ and there is an edge from $C_i$ to $A$ for $i = 1, \ldots, n$ in the network. We can clearly see that $(V, P)$ completely describes the Bayesian network.

DEFINITION 4.1. *Given a Bayesian network $B = (V, P)$, an* instantiation *is an ordered pair $(A, a)$ where $A \in V$ and $a \in R(A)$. (An instantiation $(A, a)$ is also denoted by $A = a$ and $A_a$.) A collection of instantiations $w$ is called an* instantiation-set *iff are no two instantiations $(A, a), (A, a')$ in $w$ such that $a \neq a'$.*

An instantiation represents the event when a r.v. takes on a value from its range. Given an instantiation-set, we can define the notion of the *span* of an instantiation-set.

DEFINITION 4.2. *Given an instantiation-set $w$ for a Bayesian network $B = (V, P)$, we define the* span *of $w$, $\mathrm{span}(w)$, to be the collection of r.v.s in the first coordinate of the instantiations. Furthermore, an instantiation-set $w$ is said to be* complete *iff $\mathrm{span}(w) = V$.*

NOTATION. For each r.v. $A$ and each $a$ in $R(A)$, $v_{A_a}$ is the set of all conditional probabilities in $P$ of the form $P(A = a | C_1 = c_1, \ldots, C_n = c_n)$. For each r.v. $A$, we define $\mathrm{cond}(A)$ as follows: $B \in \mathrm{cond}(A)$ iff there exists a conditional probability in $P$ of the form $P(A = a | \ldots, B = b, \ldots)$.

DEFINITION 4.3. *Given an instantiation-set $w = \{(A_1, a_1), \ldots, (A_n, a_n)\}$ for a Bayesian network $B = (V, P)$, we define the probability of $w$ to be*
$$P(w) = P(A_1 = a_1, \ldots, A_n = a_n).$$

The goal of belief revision on Bayesian networks is to determine the complete instantiation-set which maximizes the associated probability under certain conditions. In general, these conditions, called *evidence*, imposes restrictions on what instantiations may be made. The instantiation-set satisfying the evidence with the highest probability is said to be the *most probable explanation* for the evidence. We now formalize this as follows:

DEFINITION 4.4. *Given a Bayesian network $B = (V, P)$, evidence $e$ for $B$ is an instantiation-set for $B$.*

DEFINITION 4.5. *Given instantiation-sets $w_1, w_2$ for a Bayesian network $B$, $w_2$ is said to be* consistent *with $w_1$ iff $w_1 \subseteq w_2$.*

DEFINITION 4.6. *Given evidence $e$ for $B$, a complete instantiation-set $w$ for $B$ is an* explanation *for $e$ iff $w$ is consistent with $e$. Furthermore, $w$ is said to be a* most probable explanation *for $e$ iff for all explanations $w' \neq w$ for $e$, $P(w') \leq P(w)$.*

Our basic approach in constructing a constraint system from a given Bayesian network is to represent and enforce the constraints that exist between any two or more r.v.s.

---

[3] We consider prior probabilities to be degenerate cases of conditional probabilities, i.e., $P(A = a) = P(A = a | \phi)$ where $\phi$ is the empty set.

Given a Bayesian network $B = (V, P)$, we construct a constraint system $L(B) = (\Gamma, I, \psi)$ as follows:

1. For each r.v. $A$ in $V$, let $R(A) = \{a_1, \ldots, a_n\}$ and construct the variables $A_{a_1}, \ldots, A_{a_n}$ in $\Gamma$, set $\psi(A_{a_i}, \text{false}) = \psi(A_{a_i}, \text{true}) = 0$ and add the following constraint to $I$:

$$\sum_{i=1}^{n} A_{a_i} = 1. \quad (7)$$

2. For each r.v. $A$ and some $a$ in $R(A)$, for each conditional probability $P(A = a | C_1 = c_1, \ldots, C_n = c_n)$ in $v_{A_a}$, construct a variable $q[A_a | C_1 = c_1, \ldots, C_n = c_n]$ in $\Gamma$ such that (for notational convenience, we will denote $q[A_a | C_1 = c_1, \ldots, C_n = c_n]$ by $q$ in the next two conditions)

   (a) $\psi(q, \text{false}) = 0$, $\psi(q, \text{true}) = -\log(P(A = a | C_1 = c_1, \ldots, C_n = c_n))$, and,
   (b) Add the following constraint to $I$:

$$q \geq \sum_{k=1}^{n} C_{k_{C_k}} + A_a - n. \quad (8)$$

3. Let $\Upsilon_{A_a}$ be all the variables $q$ constructed by $v_{A_a}$ in step (2). For each r.v. $A$ and some $a$ in $R(A)$, add the following constraint to $I$:

$$A_a = \sum_{q \in \Upsilon_{A_a}} q. \quad (9)$$

DEFINITION 4.7. $L(B)$ constructed above is the constraint system induced by $B$.

As we can clearly see, our construction is straightforward and is done in time linear to the size of the Bayesian network. The next theorem show the complexity of our induced constraint system with respect to the Bayesian network.

THEOREM 4.1. Let $B = (V, P)$ be a Bayesian network and $L(B) = (\Gamma, I, \psi)$ be the constraint system induced by $B$. Then

1. $|\Gamma| = |P| + \sum_{A \in V} |R(A)|$ and
2. $|I| = |V| + |P| + \sum_{A \in V} |R(A)|$.

In our construction, (7) guarantees that any r.v. takes on exactly one value. (8) and (9) guarantee that the probability of any complete instantiation-set will be computed with the appropriate set of conditional probabilities. Variables of the form $q[A_a | C_{1_{C_1}}, \ldots, C_{n_{C_n}}]$ are called *conditional variables* in that they explicitly represent the dependencies between r.v.s and will be the mechanism for computing the probability for any instantiation-set.

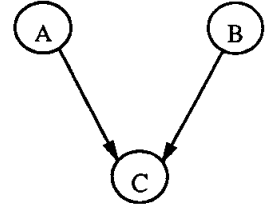

FIG. 4.1. Simple Bayesian network. The distribution is as follows:
$P(C = \text{true} | A = \text{true}, B = \text{true}) = p_1$
$P(C = \text{true} | A = \text{true}, B = \text{false}) = p_2$
$P(C = \text{true} | A = \text{false}, B = \text{true}) = p_3$
$P(C = \text{true} | A = \text{false}, B = \text{false}) = p_4$
$P(A = \text{true}) = p_9)$
$P(B = \text{true}) = p_{10})$

For example, consider the simple Bayesian network in Figure 4.1. When we have the instantiations $\{A = \text{true}, B = \text{false}, C = \text{true}\}$, its associated probability is $p_2 * p_9 * (1 - p_{10})$. In the induced constraint system, we expect our variables assignments to be $A_{\text{true}} = 1$, $B_{\text{false}} = 1$, $C_{\text{true}} = 1$, $q[C_{\text{true}} | A_{\text{true}}, B_{\text{false}}] = 1$, and all remaining variables to be 0. Since the only costs are associated with the variables $A_{\text{true}}$, $B_{\text{false}}$ and $q[C_{\text{true}} | A_{\text{true}}, B_{\text{false}}]$, the cost of this assignment is $-\log(p_9) - \log(1 - p_{10}) - \log(p_2)$ which is equivalent to $-\log(p_2 * p_9 * (1 - p_{10}))$.

NOTATION. For each r.v. $A$, let $\Delta(A)$ be the set of variables in the induced constraint system constructed for $A$.

THEOREM 4.2. Given a 0-1 solution $s$ for $L(B)$, for each set of variables $\Delta(A)$, there exists some $A_a$ in $\Delta(A)$ such that $A_a = 1$ and $A_{a'} = 0$ for all $A_{a'} \neq A_a$ in $\Delta(A)$.

THEOREM 4.3. Given a 0-1 solution $s$ for $L(B)$, for all variables $q[A_a | C_1 = c_1, \ldots, C_n = c_n]$, if $A_a = C_{1_{C_1}} = \ldots = C_{n_{C_n}} = 1$, then $q[A_a | C_1 = c_1, \ldots, C_n = c_n] = 1$.

Theorems 4.2 and 4.3 above verifies our expectations on the legitimate variable assignments. However, $A_a = C_{1_{C_1}} = \ldots = C_{n_{C_n}} = 0$ does not necessarily imply that $q[A_a | C_1 = c_1, \ldots, C_n = c_n] = 0$. We could remedy the situation by introducing the following additional constraints:

$$q[A_a | C_1 = c_1, \ldots, C_n = c_n] \leq A_a,$$

$$q[A_a | C_1 = c_1, \ldots, C_n = c_n] \leq C_{i_{C_i}} \text{ for } i = 1, \ldots, n.$$

Instead of increasing the number of constraints, we will show that this can be solved through simple restrictions and modifications to the algorithms applied to general constraint systems.

DEFINITION 4.8. A 0-1 solution $s$ for $L(B)$ is said to be permissible *if for all variables* $q[A_a | C_1 = c_1, \ldots, C_n = $

$c_n$],

$$q[\mathsf{A_a} \mid \mathsf{C_1} = c_1, \ldots, \mathsf{C_n} = c_n] = 1 \text{ only if}$$
$$\mathsf{A_a} = \mathsf{C_{1_{c_1}}} = \ldots = \mathsf{C_{n_{c_n}}} = 1.$$

Thus our goal is to consider only those 0-1 solutions for $L(\mathcal{B})$ which are permissible. We must now show that calculations on the constructed constraint system are equivalent to those on the Bayesian network for belief revision.

Given a 0-1 solution $s$ for $L(\mathcal{B})$, we can construct a complete instantiation-set $w_s$ for $\mathcal{B}$ as follows: $s(\mathsf{A_a}) = 1$ iff $(A, a) \in w_s$. To convert from a complete instantiation-set to a 0-1 solution is slightly trickier. Given a complete instantiation-set $w$ for $\mathcal{B}$, construct a 0-1 solution $s_w$ for $L(\mathcal{B})$ as follows: $(A, a) \in w$ iff $s_w(\mathsf{A_a}) = 1$. For each conditional variable $q$ in $\Upsilon_{\mathsf{A_a}}$, set the appropriate value according to $w$.

THEOREM 4.4. *If $s$ is a 0-1 solution for $L(\mathcal{B})$, then $w_s$ is an instantiation-set for $\mathcal{B}$.*

THEOREM 4.5. *If $w$ is a complete instantiation-set for $\mathcal{B}$, then $s_w$ is a permissible 0-1 solution for $L(\mathcal{B})$.*

From our construction of instantiation-sets from 0-1 solutions, we notice that more than one 0-1 solution can construct the same instantiation-set. This arises from our previous observation that our expectations are not completely met (Theorem 4.3).

COROLLARY 4.6. *There is a 1-1 and onto mapping between permissible 0-1 solutions for $L(\mathcal{B})$ and complete instantiation-sets for $\mathcal{B}$.*

This corollary states that we only need to consider the permissible 0-1 solutions in our calculations of complete instantiation-sets for the Bayesian network.

DEFINITION 4.9. *Let $e$ be some evidence for $\mathcal{B} = (V, P)$. We construct $L_e(\mathcal{B}) = (\Gamma, I_e, \psi)$ from $L(\mathcal{B}) = (\Gamma, I, \psi)$ as follows: Let $I_e = I \cup I'$ where the constraint $\mathsf{A_a} = 1$ is in $I'$ iff $(A, a) \in e$. We say that $L_e(\mathcal{B})$ is induced by $\mathcal{B}$ with evidence $e$.*

PROPOSITION 4.7. $|I_e| = |I| + |e|$.

THEOREM 4.8. *If $s$ is a 0-1 solution for $L_e(\mathcal{B})$, then $w_s$ is an explanation for $e$.*

THEOREM 4.9. *If $w$ is an explanation for $e$, then $s_w$ is a permissible 0-1 solution for $L_e(\mathcal{B})$.*

When there is some set of evidence given to be explained, we only want to consider those instantiation-sets which are consistent with the evidence. Theorems 4.8 and 4.9 above guarantee that the evidence also properly restricts the set of possible permissible 0-1 solutions we wish to consider. Now, we must show that the costs associated to each permissible 0-1 solution are directly related to the probability of the corresponding instantiation-set.

For the following theorems, assume that $L$ is induced by a Bayesian network $\mathcal{B}$, $w$ is a complete instantiation-set for $\mathcal{B}$, and $s$ is a permissible 0-1 solution for $L(\mathcal{B})$.

THEOREM 4.10. $\Theta_L(s_w) = -\log(P(w))$.

THEOREM 4.11. *There exists a constant $\alpha_e$ such that for all explanations $w$ for $e$, $\Theta_{L_e}(s_w) = \alpha_e - \log(P(w|e))$.*

THEOREM 4.12. *$w$ is a most probable explanation for $e$ iff $s_w$ is an optimal 0-1 solution for $L_e(B)$.*

Theorem 4.11 guarantees that the probabilistic ordering of instantiation-sets is exactly reversed from the cost ordering imposed on permissible 0-1 solutions. Furthermore, computing the cost for a permissible 0-1 solution immediately determines the probability of its associated instantiation-set.

THEOREM 4.13. *If $\psi(q, \text{true}) > 0$ for all conditional variables $q$ in $L_e(\mathcal{B})$, then any optimal 0-1 solution for $L_e(\mathcal{B})$ is permissible.*

The condition required in the above theorem can be easily met by increasing the cost of conditional variables with $\psi(q, \text{true}) = 0$ to $\psi(q, \text{true}) = \delta$ where $\delta$ is an arbitrarily small but positive value. This still guarantees proper ordering of the permissible 0-1 solutions as compared to the instantiation-sets.

Similarly, we must guarantee that any alternative 0-1 solutions generated must also be permissible. We can accomplish this by modifying the Algorithm 3.1. Again, instead of introducing the new constraint (6) into $L_k$ we introduce

$$\sum_{\mathsf{A_a} \in \Delta} F(s_{k_1}, \mathsf{A_a}) \leq |\Delta| - 1$$

where $\Delta = \{x | x \in V \text{ and } x \in \Delta(A) \text{ for some r.v. } A\}$.

THEOREM 4.14. *$L_n$ generates the n-th best permissible optimal 0-1 solution for $L_e(\mathcal{B})$.*

With the transformation of belief revision problems into constraint systems, we now have an alternative approach to solving for the best explanation as well as the consecutive next best. With our linear constraint satisfaction approach, we can utilize the highly efficient computational tools of operations research on the NP-Hard problem of belief revision and explanation generation. Furthermore, unlike message-passing schemes requiring preprocessing such as clustering on non-polytree topologies, our approach can be directly applied to any Bayesian network.

## 5 DISCUSSION

Linear constraint satisfaction has been shown to be an effective and computationally practical approach to solving cost-based abduction (Santos [1991a]; Santos [1991c]). Experimental results comparing our constraint system against existing search style techniques have shown it to be the superior approach.

In this paper, we have presented an approach to generating alternative explanations within our framework of constraint systems. This approach naturally incorporates the computational tools of operations research in an efficient manner. We have also shown how to apply the generation of alternative explanations to cost-based abduction and belief revision in Bayesian networks.

The necessity of having alternative explanations can also be readily seen in natural language processing. Proper handling of problems such as ambiguity requires access to the possible explanations in order of best to worst. For example, the WIMP system (Goldman [1990]; Goldman & Charniak [1991]) uses alternative explanations in order to resolve lexical ambiguities. Our approach is especially well suited to this problem since it is characterized by low prior probabilities making it monotonic within our framework.

Acknowledgments

This work has been supported by the National Science Foundation under grant IRI-8911122 and by the Office of Naval Research, under contract N00014-88-K-0589. Special thanks to Eugene Charniak for critical comments and suggestions. Also, thanks to Solomon Shimony, Glenn Carroll and Moises Lejter for careful review of this paper.

References

Charniak, Eugene & Goldman, Robert [1988], "A Logic for Semantic Interpretation," Proceedings of the AAAI Conference.

Charniak, Eugene & Shimony, Solomon E. [1990], "Probabilistic Semantics for Cost Based Abduction," Proceedings of the 1990 National Conference on Artificial Intelligence.

Genesereth, Michael R. [1984], "The Use of Design Descriptions in Automated Diagnosis," Artificial Intelligence.

Goldman, Robert P. [1990], "A Probabilistic Approach to Language Understanding," Department of Computer Science, Brown University, Ph.D. Thesis.

Goldman, Robert P. & Charniak, Eugene [1991], "Probabilistic Text Understanding," Proceedings of the Third International Workshop on AI and Statistics, Fort Lauderdale, FL.

Hobbs, Jerry R., Stickel, Mark, Martin, Paul & Edwards, Douglas [1988], "Interpretation as Abduction," Proceedings of the 26th Annual Meeting of the Association for Computational Linguistics.

Kautz, Henry A. & Allen, James F. [1986], "Generalized Plan Recognition," Proceedings of the Fifth Conference of AAAI.

Nemhauser, G. L., Kan, A. H. G. Rinnooy & Todd, M. J. [1989], in *Optimization: Handbooks in Operations Research and Management Science Volume 1*, North Holland.

Pearl, Judea [1988], in *Probabilistic Reasoning in Intelligent Systems: Networks of Plausible Inference*, Morgan Kaufmann, San Mateo, CA.

Peng, Y. & Reggia, J. A. [1990], in *Abductive Inference Models for Diagnostic Problem-Solving*, Springer-Verlag.

Santos, Eugene Jr. [1991a], "A Linear Constraint Satisfaction Approach to Cost-Based Abduction," Department of Computer Science, Brown University, in preparation.

Santos, Eugene Jr. [1991b], "Cost-Based Abduction, Linear Constraint Satisfaction, and Alternative Explanations," to appear in Proceedings of the AAAI Workshop on Abduction.

Santos, Eugene Jr. [1991c], "Cost-Based Abduction and Linear Constraint Satisfaction," Department of Computer Science, Brown University, Technical Report CS-91-13.

Selman, Bart & Levesque, Hector J. [1990], "Abductive and Default Reasoning: A Computational Core," Proceedings of the Eighth National Conference on Artificial Intelligence.

Shanahan, Murray [1989], "Prediction is Deduction but Explanation is Abduction," IJCAI-89.

Stickel, Mark E. [1988], "A Prolog-like Inference System for Computing Minimum-Cost Abductive Explanations in Natural-Language Interpretation," SRI International, Technical Note 451.